\documentclass[10pt,twocolumn,letterpaper]{article}

\usepackage{cvpr}              
\usepackage{url}
\usepackage{amsmath}
\usepackage{multirow}
\usepackage{array}
\usepackage{booktabs}
\usepackage{graphicx} 
\usepackage{caption} 
\usepackage{lscape} 
\usepackage{bbding}
\usepackage{wrapfig}
\usepackage{color}
\usepackage{xcolor}

\definecolor{myblue}{HTML}{6093D4}
\definecolor{mypurple}{HTML}{CA74C7}
\definecolor{mygreen}{HTML}{65A542}

\definecolor{cvprblue}{rgb}{0.21,0.49,0.74}
\usepackage[pagebackref,breaklinks,colorlinks,allcolors=cvprblue]{hyperref}

\def\paperID{13829} 
\def\confName{CVPR}
\def\confYear{2025}

\title{Compositional Image Retrieval via Instruction-Aware Contrastive Learning}

\author{Wenliang Zhong, Weizhi An, Feng Jiang, Hehuan Ma, Yuzhi Guo, and Junzhou Huang\\
UT Arlington\\
{\tt\small \{wxz9204, weizhi.an, fxj8843, hehuan.ma, yuzhi.guo\}@mavs.uta.edu}\\
{\tt\small jzhuang@uta.edu}
}

\begin{document}
\maketitle
\begin{abstract}
Composed Image Retrieval (CIR) involves retrieving a target image based on a composed query of an image paired with text that specifies modifications or changes to the visual reference. CIR is inherently an instruction-following task, as the model needs to interpret and apply modifications to the image. 
In practice, due to the scarcity of annotated data in downstream tasks, Zero-Shot CIR (ZS-CIR) is desirable.
While existing ZS-CIR models based on CLIP have shown promising results, their capability in interpreting and following modification instructions remains limited. Some research attempts to address this by incorporating Large Language Models (LLMs). However, these approaches still face challenges in effectively integrating multimodal information and instruction understanding.
To tackle above challenges, we propose a novel embedding method utilizing an instruction-tuned Multimodal LLM (MLLM) to generate composed representation, which significantly enhance the instruction following capability for a comprehensive integration between images and instructions. 
%
Nevertheless, directly applying MLLMs introduces a new challenge since MLLMs are primarily designed for text generation rather than embedding extraction as required in CIR. To address this, we introduce a two-stage training strategy to efficiently learn a joint multimodal embedding space and further refining the ability to follow modification instructions by tuning the model in a triplet dataset similar to the CIR format. 
Extensive experiments on four public datasets: FashionIQ, CIRR, GeneCIS, and CIRCO demonstrates the superior performance of our model,  outperforming state-of-the-art baselines by a significant margin.
Codes are available at the \href{https://github.com/zhwl2117/InstructCIR.git}{GitHub repository}.
\end{abstract}


\section{Introduction}
Composed Image Retrieval (CIR) refers to retrieving a target image based on a composed query consisting of both an image and accompanying text~\citep{saito2023pic2word, gulanguage, agnolucci2024isearle}. The textual input typically serves as a modification instruction applied to the visual reference, guiding the retrieval process. Such tasks are prevalent in practical applications, particularly in e-commerce scenarios~\citep{barbany2024leveraging, zhu2024bringing}, where users might wish to find visually similar items with slight difference, such as a change in color or style. However, unlike conventional image-text retrieval tasks, CIR presents unique challenges in data acquisition for different downstream tasks, as it necessitates the creation of triplet data \textit{$($}\textit{source image, modifier text, target image}\textit{$)$}. This requirement significantly increases the complexity and cost of data collection, as human annotators are often needed to generate specific textual descriptions that link relevant images. 
To address the limitations, recent research~\citep{baldrati2023zero, ventura2024covr} has focused on Zero-Shot CIR (ZS-CIR) as a scalable approach, which are trained on a large-scale dataset and can be directly applied to diverse contexts. 
\begin{figure}[t]
    \centering
    \includegraphics[width=1.0\linewidth]{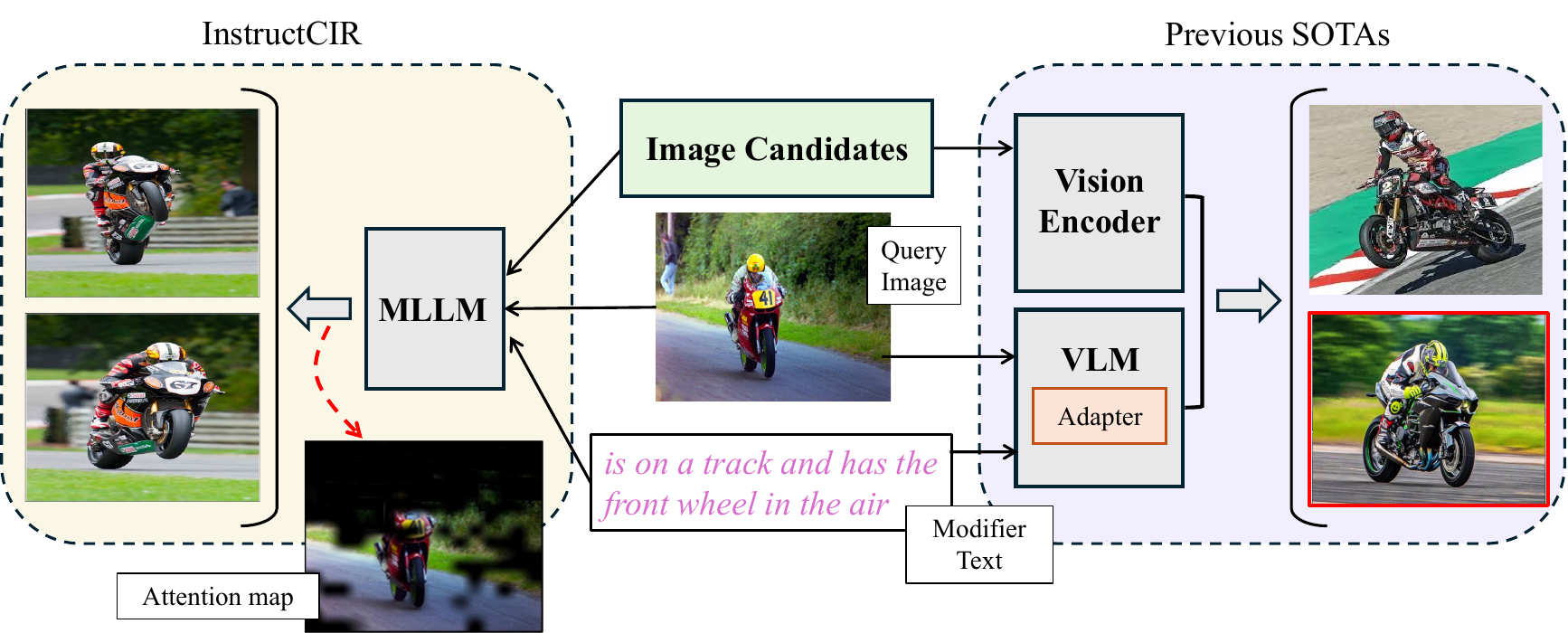}
    \caption{\textbf{Comparison of Existing ZS-CIR Approaches vs. InstructCIR.} Current state-of-the-art CIR methods typically rely on VLMs such as CLIP. These methods are constrained by the limited instruction-following capabilities in CLIP models. In contrast, Our approach employs instruction-tuned MLLMs specifically designed for instruction-following tasks including CIR. As shown in the attention map derived from the composed embedding using \cite{yu2024attention}. Our approach is able to focus on specific parts of the image following the modification instruction. In the example, the front wheel and the floor are highlighted according to the “on a track” and “front wheel in the air” of the modification.}  
    \label{fig: overview}
\end{figure}

Most existing ZS-CIR models build on CLIP-based architectures~\citep{radford2021learning}, leveraging their robust visual-text representation capabilities. For example, Pic2Word~\citep{saito2023pic2word} and SEARLE~\citep{agnolucci2024isearle} utilize lightweight projection modules to map visual embeddings into the textual space, enhancing the interaction between visual and textual modalities within CLIP's framework. Similarly, LinCIR~\citep{gulanguage} introduces a language-only training strategy, utilizing keywords in text to represent images. 
While these methods are effective, they are fundamentally constrained by the lack of the instruction-following capability within CLIP models~\citep{wei2023uniirtrainingbenchmarkinguniversal}. Nevertheless, Composed Image Retrieval is inherently an instruction-following task because the model needs to comprehend the modification and applied it to the image. For example, when the modification is "changing the dog to a cat", the model should generate a composed embedding containing the basic image information but replacing the dog semantic to a cat. Unfortunately, existing CLIP-based models fail to provide a comprehensive composed embedding for the image and modification instruction. 

Recent works try to incorporate the instruction understanding capability in Large Language Models (LLMs)~\citep{zhao2023survey} to tackle this challenge. For instance, CIReVL~\citep{karthik2023vision} leverages ChatGPT~\cite{brown2020language} to combine image captions and textual instructions, thereby enabling a training-free retrieval process. However, the involvement of ChatGPT may be prohibited in commercial scenarios due to privacy concerns. The image caption generated in inference can also be inaccurate. VDG~\citep{jang2024visual} proposes generating triplet data using a trained Multimodal LLMs (MLLMs)~\citep{yin2023survey}, but the MLLM itself remains peripheral to the retrieval process, limiting its direct impact on model performance. Approaches such as FROMAGe~\citep{koh2023grounding} and MCL~\citep{liimproving} employ image captioning and contrastive learning to integrate LLMs with visual encoders, yet they freeze the LLMs to function only as static encoders. Consequently, these models do not fully exploit the adaptability and instruction awareness that LLMs can offer for more detailed query comprehension in ZS-CIR tasks because of the indirect application of LLMs. 

To tackle the lack of instruction-following capability in CLIP-based models and fully exploit the instruction understanding capability of MLLMs for CIR, we introduce a novel embedding method using pure instruction-tuned MLLMs, which offers two key advantages. First, a solid vision-text alignment is provided which is crucial for multimodal tasks like CIR. Second, MLLMs are designed to follow complex instructions, a capability learned during instruction tuning. However, despite these potentials, MLLMs have been primarily used for text generation tasks, and their application to CIR has not been thoroughly explored. 
To induce the capability of MLLMs in ZS-CIR thoroughly, we introduce a two-stage training strategy to adapt MLLMs for CIR. In the first stage, we perform contrastive learning~\citep{chen2020simple} using pure image-text pairs to shift the MLLM's function from text generation to representation derivation, enabling it to produce multimodal embeddings suitable for retrieval. However, the pair-wise retrieval still deviates from CIR because the composed embedding requires both the image and instruction. 
Hence, in the second stage, we enhance the MLLM’s instruction-awareness by tuning it on a triplet dataset similar to CIR tasks. Specifically, the MLLM is tuned to produce  embeddings based on the composition of the image and modification instruction to align with the target caption embedding.
Our approach, named \textbf{InstructCIR}, significantly enhances model performance on ZS-CIR benchmarks as shown in Figure \ref{fig: overview}.

In summary, our contributions are fourfold: 
(1) We delineate that Composed Image Retrieval is an instruction-following task and existing ZS-CIR models based on CLIP lack such a capability. Conversely, we propose an embedding strategy based on instruction-tuned MLLMs, providing superior instruction-following capabilities over previous approaches.
(2) To mitigate the task discrepancy between the image-text retrieval and CIR, we construct a triplet dataset similar to the CIR format, serving as an ideal training resource for aligning the composed source embedding and target embedding.
(3) To fully harness the capabilities of MLLMs for composed image retrieval, we introduce a two-stage training strategy that not only adapts the MLLM’s strong text generation capabilities to effective representation derivation but also optimally fine-tunes it for ZS-CIR tasks.
(4) Extensive experiments are conducted on four public datasets: FashionIQ, CIRR, GeneCIS, and CIRCO. Results reveal the superiority of InstructCIR, outperforming existing state-of-the-arts baselines by a significant margin.
\begin{figure*}[t]
    \centering
    \includegraphics[width=1.0\linewidth]{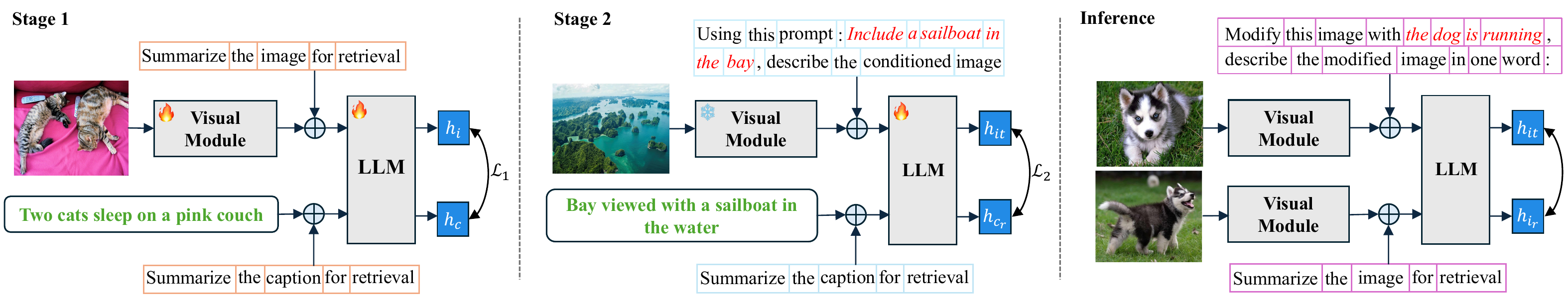}
\caption{
\textbf{The Two-Stage Training Strategy for InstructCIR.} 
The diagram illustrates our two-stage approach. \textbf{Stage 1}: The model is trained on image-caption pairs $(i, c)$ to align multimodal embeddings. The image is encoded by the MLLM to $h_i$, while the caption is processed to generate $h_c$. This stage establishes a shared embedding space for both modalities. \textbf{Stage 2}: The model is fine-tuned with triplet data $(i, t, c_r)$. The image and modifier text are composed into an embedding $h_{it}$, while the modified caption is encoded as $h_{c_r}$. The objective is to align $h_{it}$ and $h_{c_r}$, enhancing instruction-following abilities. The visual module includes the visual encoder and adapter. The strategy effectively handles CIR tasks by integrating visual and textual information. \textbf{Inference}: During inference, the source image is encoded with the corresponding modification instruction to $h_{it}$. Target images are encoded to $h_{i_r}$, which can be pre-computed and cached. The CIR system leverages the composed embedding $h_{it}$ to find the matched target image embedding $h_{i_r}$.
}

\label{fig: two_stages}
\end{figure*}

\section{Methodology}
In this section, we first outline the preliminaries of CIR and introduce the notations used in this paper. We then present InstructCIR, an MLLM-based embedding model capable of processing images, text, or a combination of both to generate a unified embedding. The embedding captures a comprehensive composition of reference images and textual instructions. To train this model, we propose a two-stage training strategy as shown in Figure \ref{fig: two_stages}. The first stage focuses on creating a joint embedding space, where we utilize pure image-text pairs to train the MLLM as an effective embedding model. This step is crucial for transitioning the MLLM from a text generation role to that of representation derivation. In the second stage, we train the model to produce instruction-aware composed embeddings. Specifically, a triplet dataset \textit{$($}\textit{source image, modifier text, target caption}\textit{$)$} similar to the CIR format is constructed with the help of GPT-4o to generate altering instructions and corresponding modified captions. The model is then trained to align the image-instruction embeddings with the target caption embeddings. The two-stage framework allows the MLLM to learn both modality alignment and instruction-following capabilities, which are essential for effective CIR.


\subsection{Preliminary}

CIR involves retrieving target images based on a combination of a reference image and a modifier text which we term \textit{instruction} or \textit{prompt}. Formally, given a reference image $i \in \mathcal{I}$ and an instruction $t \in \mathcal{T}$ that describes the desired modification, the composed query $(i, t)$ is used to search for the closest target image $i_r$ within an image database. The primary challenge in CIR lies in generating unified embeddings that can effectively represent the composition of both visual and instructional information to match ideal target image embeddings. InstructCIR is able to follow the instruction to create a composed embedding. The inference stage of InstructCIR is shown in Figure \ref{fig: two_stages}.


\subsection{Constructing the Instruction-Aware Dataset}

There is a task discrepancy between the image-text retrieval and composed image retrieval~\citep{byun2024reducing}, hindering previous training strategies relying on the pair data. Consequently, a triplet dataset similar to the CIR format is an optimal resource to align composed embeddings. In this subsection, we outline the process of constructing such a dataset. Drawing inspiration from MCL~\citep{liimproving}, we induce triplet data from the pair data available in the existing dataset CC3M~\citep{sharma-etal-2018-conceptual}. Specifically, for each image-caption pair $(i, c)$, we utilize the caption $c$ to represent the image $i$. 
Unlike MCL, we propose leveraging GPT-4o~\citep{achiam2023gpt} using the Chain of Thought method~\citep{wei2022chain}. This involves providing GPT with the original caption $c$ and few-shot examples. GPT is asked to brainstorm a triplet step by step. Specifically, it firstly identify key concepts in the caption, then derive a change to specific concepts as a modification instruction. Finally, the modified caption is generated based on the source caption and modification instruction. For instance, given the caption “A husky is lying on the grass,” we identify the object husky, the action lying, and the background grass. By changing the action to running, the modified caption becomes “A husky is running on the grass.” The query caption $c$ is given to GPT followed by examples, resulting in a triplet $(i, t, c_r)$ where $t$ is the brainstormed instruction and $c_r$ is the modified caption. Due to space limitations, figures illustrating the processing pipeline and difference from the MCL data processing are included in Appendix \ref{sec: triplet_data}. It is worthy mentioning that though existing methods~\cite{karthik2023vision} have demonstrated that the modified caption generated by ChatGPT can be directly used in inference, it is not always feasible because ChatGPT may be prohibited for privacy reasons in commercial scenarios. In addition, the inaccurate caption can directly hamper the final retrieval. 

Notably, acquiring the modified image $i_r$ is often more complex and costly. Therefore, we use the constructed triplet $(i, t, c_r)$ directly for model training. Since the ultimate retrieval target in CIR is an image, rather than text, our first training stage learns an aligned embedding space. This alignment ensures that, when the model is trained to retrieve the modified caption in the second stage, the resulting embeddings are consistent with those of the modified images. This approach facilitates effective training for CIR by aligning textual modifications with visual changes.
\begin{figure}[!h]
    \centering
    \includegraphics[width=0.32\textwidth]{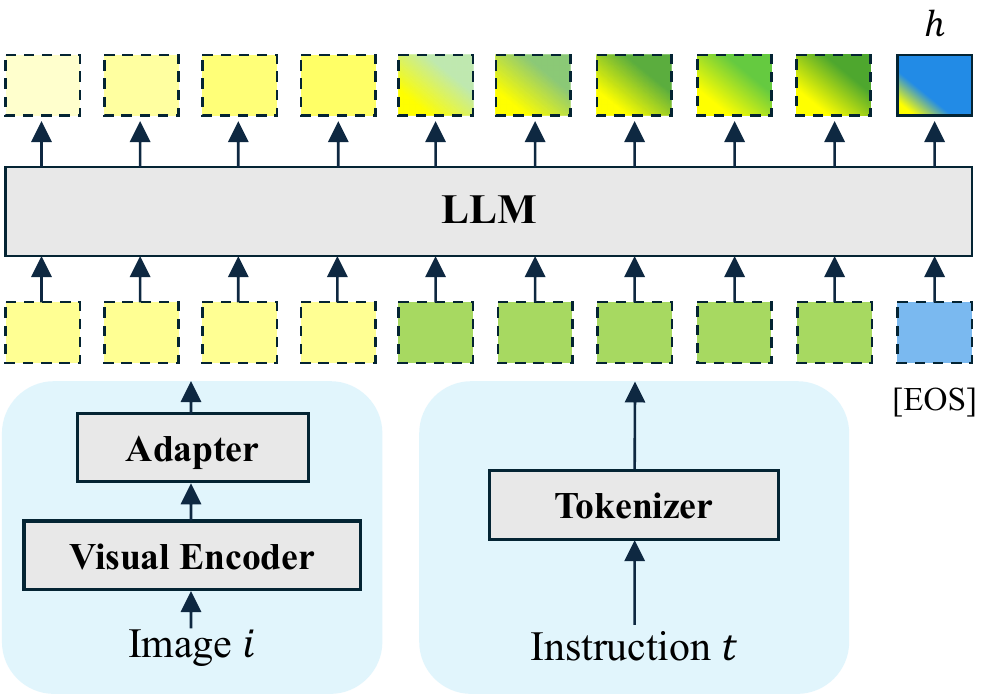}
    \caption{
    \textbf{Model Architecture}: For composed inputs (images and texts), the image $i$ is processed by a visual encoder and adapter, while the instruction $t$ is tokenized. Both are concatenated and fed into the LLM along with the [EOS] token. The final output at the [EOS] token provides the unified embedding $h$. For text-only inputs, the visual encoder and adapter are bypassed. The Causal Attention in the LLM update previous token information into the current token, comprehensively integrating the image and instruction information into the [EOS] and finally resulting in an instruction-aware composed embedding $h$.
    }
    \label{fig: model_arch}
\end{figure}

\subsection{Instruction-Aware Contrastive Learning}
\textbf{Model Architecture.} To fully leverage the instruction following capability, we propose to use an instruction-tuned MLLM for embedding. In common MLLMs, images are processed by the visual encoder, such as a Vision Transformer (ViT)~\citep{alexey2020image}. The resulting patch embeddings are then projected into the LLM embedding space via an adapter, allowing them to be concatenated with the input text embeddings. 
The concatenated sequence is subsequently fed into the LLM component to produce the final output. When only textual input, such as captions and prompts, is provided, it is directly tokenized and processed by the LLM, bypassing the visual encoder. To extract a comprehensive embedding from the MLLM, we append a special token [EOS] at the end of the input sequence. The input sequence, including this [EOS] token, is forwarded to the model, and the embedding corresponding to the [EOS] token in the output is used as the global representation $h$. This forward process leverages the Causal Attention mechanism within the LLM, where the current token will aggregate previous token information and the final [EOS] token will include the entire information of the image and instruction information in a causal manner, making it instruction-aware. The model architecture is illustrated in Figure \ref{fig: model_arch}. We use subscripts to denote representations from different inputs in later sections.

\begin{equation}
\begin{aligned}
    \mathcal{L}_{i2c} &= -\text{log} (\frac{\phi (h_i, h_c)}{\phi (h_i, h_c) + \sum\limits_{n \in \mathbb{N}_1} \phi (h_i, h_{n_c})})\\
    \mathcal{L}_{c2i} &= -\text{log} (\frac{\phi (h_c, h_i)}{\phi (h_c, h_i) + \sum\limits_{n \in \mathbb{N}_1} \phi (h_c, h_{n_i})})\\
    \mathcal{L}_1 &= \frac{1}{2} \left(\mathcal{L}_{i2c} + \mathcal{L}_{c2i}\right)
\end{aligned}
\label{equ: stage1_loss}
\end{equation}

\textbf{Learning Retrieval Embeddings.} There is a task discrepancy between text generation and embedding extraction for MLLMs. In the first stage, we aim to learn a joint multimodal embedding space for retrieval. Specifically, we leverage image-caption data $(i, c) \in \mathcal{D}_1$ for contrastive learning. Inspired by \cite{jiang2023scaling}, an instruction “\textit{Summarize the image (caption) in one word:}” is used to prompt the model for summarizing each image or text. Both the image and the text with their instructions are then fed into the model to obtain the embeddings $(h_i, h_c)$. The model is trained using an InfoNCE loss~\citep{oord2018representation}, as in Equation \ref{equ: stage1_loss}. During this stage, all components of the MLLM are trainable, aiming to facilitate the learning of a unified embedding space.

Here, $\phi (h_i, h_c) = \text{exp} \left(\frac{1}{\tau} \text{cos} (h_i, h_c)\right)$ represents the scaled cosine similarity, where $\tau$ is the temperature parameter. $\mathbb{N}_1$ denotes the set of negative samples for the current batch, and $h_{n_i}$ ($h_{n_c}$) refers to negative image (caption) correspondences. We utilize in-batch samples as well as hard negative samples (if provided) to construct the negative set. Details are shown in the \textbf{left} part of Figure \ref{fig: two_stages}.

\textbf{Instruction Contrastive Tuning.} From the previous stage, we train an MLLM-based embedding model encoding multimodal inputs into a joint embedding space. However, the embedding model is not suitable for CIR as it is not responsive to different modification instructions. In this stage, we address this problem by tuning the model in the triplet data similar to the CIR format. In specific, we incorporate different instructions and images as composed embeddings and align them with corresponding target captions. The training strategy generalizes the instruction awareness to unseen CIR scenarios. 

Given the generated triplet data $(i, t, c_r) \in \mathcal{D}_2$, we use a prompt template to integrate the modification instruction $t$. This template, such as \textit{“Using this prompt: \{\}, describe the conditioned image: ”}, is sampled from a predefined set and is designed to guide the model in understanding how the image should be modified according to the instruction. The reference image $i$ and the formatted instruction are encoded by the model into a composed embedding $h_{it}$.

Another template is employed to guide the model in retrieving the modified caption. We use a summary prompt, such as \textit{“Summarize the caption for retrieval: ”}, sampled from another predefined set, to encode the modified caption $c_r$. This template helps the model learn to distill key information into a retrieval-friendly representation. The model encodes the prompt and the modified caption to generate the embedding $h_{c_r}$. Details of the prompt template sets are provided in Appendix \ref{sec: prompt_tempalate}.

By using different prompts, we encourage the model to distinguish the task of understanding modification instructions and the task of summary for retrieval. This distinction is crucial for enhancing the model's ability to generalize to unseen data in a zero-shot setting. Finally, we compute the InfoNCE loss between the composed embedding $h_{it}$ and the target embedding $h_{c_r}$ as shown in Equation \ref{equ: stage2_loss}. During this stage, the visual encoder and adapter are frozen, and only the LLM is trained to refine its instruction-following capabilities. Details of this stage is shown in the \textbf{middle} part of Figure \ref{fig: two_stages}.

\begin{equation}
    \mathcal{L}_2 = - \text{log} (\frac{\phi (h_{it}, h_{c_r})}{\phi (h_{it}, h_{c_r}) + \sum\limits_{n \in \mathbb{N}_2} \phi (h_{it}, h_n)})
    \label{equ: stage2_loss}
\end{equation}

Here, $\phi$ represents the scaled cosine similarity. The negative set $\mathbb{N}_2$ consists of other in-batch modified captions and the original caption $c$ of the current sample.

\section{Experiments}
\subsection{Settings}
\begin{table*}[!t]
\scriptsize
\centering
\begin{tabular}{lcccccccccc}
\toprule
\multirow{3}{*}{Method} & \multicolumn{6}{c}{\textbf{CIRR}} & \multicolumn{4}{c}{\textbf{CIRCO}} \\ 
\cmidrule(lr){2-7} \cmidrule(lr){8-11}
 & $R@1$   & $R@5$   & $R@10$ & $R_s @1$ & $R_s @2$ & $R_s @3$ & $mAP@5$ & $mAP@10$ & $mAP@25$ & $mAP@50$ \\
\midrule
Pic2word$^{\dag}$~\citep{saito2023pic2word}    & 23.90  & 51.70  & 65.30 & 53.76 & 74.46 & 87.08 & 8.72  & 9.51   & 10.64  & 11.29 \\
SEARLE$^{\dag}$~\citep{baldrati2023zero}      & 24.24 & 52.48 & 66.29 & 53.76 & 75.01 & 88.19 & 11.68 & 12.73  & 14.33  & 15.12 \\
KEDs$^{\dag}$~\citep{suo2024knowledge}        & 26.40  & 54.80  & 67.20 & - & - & - & -     & -      & -      & -     \\
Context-I2W$^{\dag}$~\citep{tang2024context} & 25.60  & 55.10  & 68.50 & - & - & - & -     & -      & -      & -     \\
LinCIR$^{\dag}$~\citep{gulanguage}      & 25.04 & 53.25 & 66.68 & 57.11 & 77.37 & 88.89 & 12.59 & 13.58  & 15.00     & 15.85 \\
\midrule
Image2Sentence$^{\ddag}$~\citep{du2024image2sentence} & 30.84 & 61.06 & 73.57 & - & - & - & 13.33 & 12.25 & 13.42 & 13.97 \\
Slerp$^{\ddag}$~\citep{jang2024spherical} & 28.60 & 55.37 & 65.66 & 65.16 & 83.90 & 92.05 & 9.61 & 10.11 & 11.10 & 11.66 \\
\midrule
CIReVL$^{*}$~\citep{karthik2023vision}      & 24.55 & 52.31 & 64.92 & 59.54 & 79.88 & 89.69 & 18.57 & 19.01  & 20.89  & 21.80  \\
FROMAGe$^{*}$~\citep{koh2023grounding}     & 10.96 & 31.40  & - & 34.07 & - & - & 4.00 & 4.44   & 5.26   & 5.73  \\
MCL$^{*}$\citep{liimproving}         & 26.22 & 56.84 & - & 61.45 & - & - & 17.67 & 18.86  & 20.80   & 21.68 \\
\midrule
\textbf{InstructCIR$_{lp}$} & \underline{35.08} & \textbf{65.25} & \underline{76.53} & \underline{67.52} & \underline{84.13} & \underline{92.08} & \underline{22.19} & \underline{23.62} & \underline{26.01} & \underline{27.20} \\
\textbf{InstructCIR$_{full}$} & \textbf{35.18} & \underline{65.12} & \textbf{77.61} & \textbf{67.54} & \textbf{84.77} & \textbf{93.61} & \textbf{22.32} & \textbf{23.80} & \textbf{26.25} & \textbf{27.32}\\
\bottomrule
\end{tabular}
\caption{\textbf{Comparison of Zero-Shot CIR Models on CIRCO and CIRR Test Sets.}  Baseline results are directly taken from original papers. “\dag” represents \textbf{CLIP-based models}; “\ddag” represents \textbf{BLIP-based models}; and “*” represents \textbf{LLM-based models}. $full$ indicates the model trained with LLaVA-Pretrain and FOIL while $lp$ indicates the model trained with LLaVA-Pretrain only in the first stage. \textbf{Bold} indicates the highest score and \underline{Underline} indicates the second highest. Results not reported are marked as “-”. Our model significantly outperforms baseline ZS-CIR models across various metrics and datasets.}
\label{tab:cirr_and_circo_results}
\end{table*}

For our experiments, we adopt the xtuner/llava-phi-3-mini-hf~\citep{2023xtuner} as the base model for InstructCIR, chosen for two key reasons: (1) LLaVA-based models~\citep{liu2024visual} represent a widely-used paradigm in current MLLMs, and testing on such a model provides valuable insights that can be generalized to similar architectures. (2) On-device LLMs such as Phi-3-mini\citep{abdin2024phi} are much smaller than general LLMs, enabling them to be runnable even in mobile devices and inferred much faster. To ensure consistency with the baseline models, we do not directly apply the checkpoint from xtuner/llava-phi-3-mini-hf. Instead, we re-train a variant, denoted as LLaVA-Phi, by modifying the visual encoder from the original openai/clip-vit-large-patch14-336 (which uses a $336\times 336$ resolution) to openai/clip-vit-large-patch14 with a $224\times 224$ resolution. Additionally, we replace the LLM component to the latest Phi-3.5-mini. 
In ablation studies, we show our training strategy can be directly applied to existing MLLMs, such as microsoft/Phi-3.5-vision-instruct~\citep{abdin2024phi}, and explore cutting-edge techniques like dynamic high-resolution for CIR tasks.

For the first stage of training, we utilize two image-caption datasets: LLaVA-Pretrain~\citep{liu2024llava} and FOIL~\citep{shekhar2017foil_acl}, an extension of the MSCOCO 2014 dataset~\citep{lin2014microsoft}, where each image-caption pair includes hard negative captions to enhance learning. The reason we leverage two datasets is to demonstrate that the training can be benefited from more diverse context as delineated in the ablation study. In the second stage, we derive a 2M triplet dataset from the CC3M, termed \textbf{CC3M-Instruct}. The more data details are provided in Appendix \ref{sec: data_stats}. We randomly select a 300K subset from CC3M-Instruct, as it provides efficient training without loss in performance compared to the full dataset. 
\textbf{Importantly}, some benchmark datasets also utilize images from MSCOCO. However, they are sourced from different versions and dataset splits with FOIL. Moreover, aside from the images, FOIL does not include the modification instructions and corresponding targets from these benchmarks but contains only captions. Therefore, there is no overlap between our training and testing settings. To further eliminate the concern in a strict zero-shot setting, we report two results: InstructCIR$_{lp}$ trained with LLaVA-Pretrain only in the first stage and InstructCIR$_{full}$ trained with both LLaVA-Pretrain and FOIL in the first stage.
Both stages are trained for one epoch. To optimize efficiency, we employ LoRA~\citep{hu2021lora} and DeepSpeed ZeRO-2~\citep{rajbhandari2020zero} during training. The model is trained on a cluster of four H100 GPUs. More hyperparameters and configuration details are included in Appendix \ref{sec: train_details}. All codes, processed datasets, and model checkpoints will be released to the public to ensure reproducibility.
\begin{table}[!ht]
\scriptsize
\centering
\scalebox{0.89}{
\begin{tabular}{lcccccc}
\toprule
\multirow{3}{*}{Method} & \multicolumn{2}{c}{\textbf{Shirt}} & \multicolumn{2}{c}{\textbf{Dress}} & \multicolumn{2}{c}{\textbf{Toptee}} \\ 
\cmidrule(lr){2-3} \cmidrule(lr){4-5} \cmidrule(lr){6-7}
& $R@10$ & $R@50$ & $R@10$ & $R@50$ & $R@10$ & $R@50$\\ 
\midrule
Pic2word~\citep{saito2023pic2word} & 26.20 & 43.60 & 20.00 & 40.20 & 27.90 & 47.40 \\ 
SEARLE~\citep{baldrati2023zero} & 26.89 & 45.58 & 20.48 & 43.13 & 29.32 & 49.97 \\ 
KEDs~\citep{suo2024knowledge}  & 28.90 & 48.00 & 21.70 & 43.80 & 29.90 & 51.90 \\ 
Context-I2W~\citep{tang2024context} & 29.70 & 48.60 & 23.10 & 45.30 & 30.60 & 52.90 \\ 
LinCIR~\citep{gulanguage} & 29.10 & 46.81 & 20.92 & 42.44 & 28.81 & 50.18 \\ 
Image2Sentence~\citep{du2024image2sentence} & 30.03 & 48.58 & 25.33 & 46.26 & 33.45 & 53.80\\
Slerp~\citep{jang2024spherical} & 27.33 & 45.25 & 22.91 & 42.39 & 32.33 & 50.48\\
CIReVL~\citep{karthik2023vision} & 29.49 & 47.40 & 24.79 & 44.76 & 31.36 & 53.65 \\ 
\midrule
\textbf{InstructCIR$_{lp}$} & \underline{31.38} & \underline{51.42} & \underline{27.06} & \underline{48.56} & \underline{36.75} & \underline{55.77} \\
\textbf{InstructCIR$_{full}$} & \textbf{32.24} & \textbf{52.11} & \textbf{28.15} & \textbf{49.38} & \textbf{37.26} & \textbf{56.13} \\ 
\bottomrule
\end{tabular}%
}
\caption{\textbf{Comparison of Zero-Shot CIR Models on FashionIQ}. }
\label{tab: fiq_results}
\end{table}

\subsection{Datasets and Baselines}
We evaluate our model using four well-established zero-shot CIR benchmarks: FashionIQ~\citep{guo2019fashion}, CIRR~\citep{Liu_2021_ICCV}, CIRCO~\citep{baldrati2023zeroshot}, and GeneCIS~\citep{vaze2023genecis}. More details of datasets are in Appendix \ref{sec: data_detail}.
In line with common practice, we report Recall@k ($R@k$) for FashionIQ, CIRR, and GeneCIS, with an additional subset metric for CIRR denoted as $R_s@k$. For CIRCO, where multiple correct images can correspond to a single query, we use mean Average Precision ($mAP@k$) to capture both precision and recall across different retrieval positions. Note that CIRR and CIRCO have hidden test sets accessible only through server submissions. We report the main results on these test sets following baseline protocols but conduct ablations on the corresponding validation sets except Section \ref{sec: advanced_mllm}.

We compare our approach against state-of-the-art CIR models, focusing on those that use ViT-L ($224\times224$) as the visual backbone. These baselines are divided into three categories: (1) CLIP-based models, including Pic2Word~\citep{saito2023pic2word}, Context-I2W~\citep{tang2024context}, KEDs~\citep{suo2024knowledge}, SEARLE~\citep{baldrati2023zero}, and LinCIR~\citep{gulanguage}; (2) BLIP~\citep{li2022blip}-based models, containing Image2Sentence~\citep{du2024image2sentence} and Slerp~\citep{jang2024spherical}; (3) LLM-based models, such as FROMAGe~\citep{koh2023grounding}, CIReVL~\citep{karthik2023vision}, and MCL~\citep{liimproving}. We exclude baselines~\citep{zhang2024magiclens, gu2023compodiff} trained on unreleased data, as their data distribution is unknown. Notably, CIReVL~\cite{karthik2023vision} conducts ZS-CIR with an image captioner and ChatGPT to combine image captions and modifications to modified captions, and then uses it with a retrieval model for target images. The process of generating modified captions is similar to our triplet data generation and can serve as a baseline about directly using ChatGPT for ZS-CIR.


\subsection{Main Results}
For the CIRCO benchmark, Table \ref{tab:cirr_and_circo_results} reports performance on the hidden test set, which is accessible via the submission server provided by \cite{baldrati2023zero}. Our approach demonstrates substantial improvements over existing methods, such as Pic2Word and SEARLE, achieving an $mAP@5$ of 22.32\%. This represents a notable increase of 13.60\% over Pic2Word and 10.64\% over SEARLE. Additionally, when compared to CIReVL, which leverages BLIP-2 and ChatGPT, our model achieves an improvement of 4.79\% in $mAP@5$. These results are particularly significant given that CIRCO is the most rigorously annotated dataset in the CIR field. Unlike other datasets, CIRCO incorporates multiple correct target images for each query, addressing the inherent ambiguity of the CIR task, where textual modifications of an image can yield multiple valid outcomes. The strong performance of our model on this dataset provides key evidence of its robustness and ability to handle complex retrieval tasks with greater precision than current state-of-the-art methods.

For the CIRR dataset, the results from the hidden test set, returned by the submission server as in \cite{liu2021image}, are also in Table \ref{tab:cirr_and_circo_results}. CIRR presents unique challenges due to its noisy nature, where the modifying instruction plays a much larger role in the retrieval process, while the reference image often has less direct correlation with the target image. Despite this noise, our model achieves substantial improvements, surpassing Pic2Word and SEARLE by 11.28\% and 10.94\% in $R@1$, respectively. Among the baselines, the most competitive result comes from MCL, an LLM-based model also trained on triplet data. However, our model surpasses MCL by 8.96\% in $R@1$ and 6.09\% in $R_s@1$, underscoring the effectiveness and flexibility of our approach in handling complex CIR tasks where the relationship between images and instructions is ambiguous.
\begin{figure*}[t]
    \centering
    \includegraphics[width=1.0\linewidth]{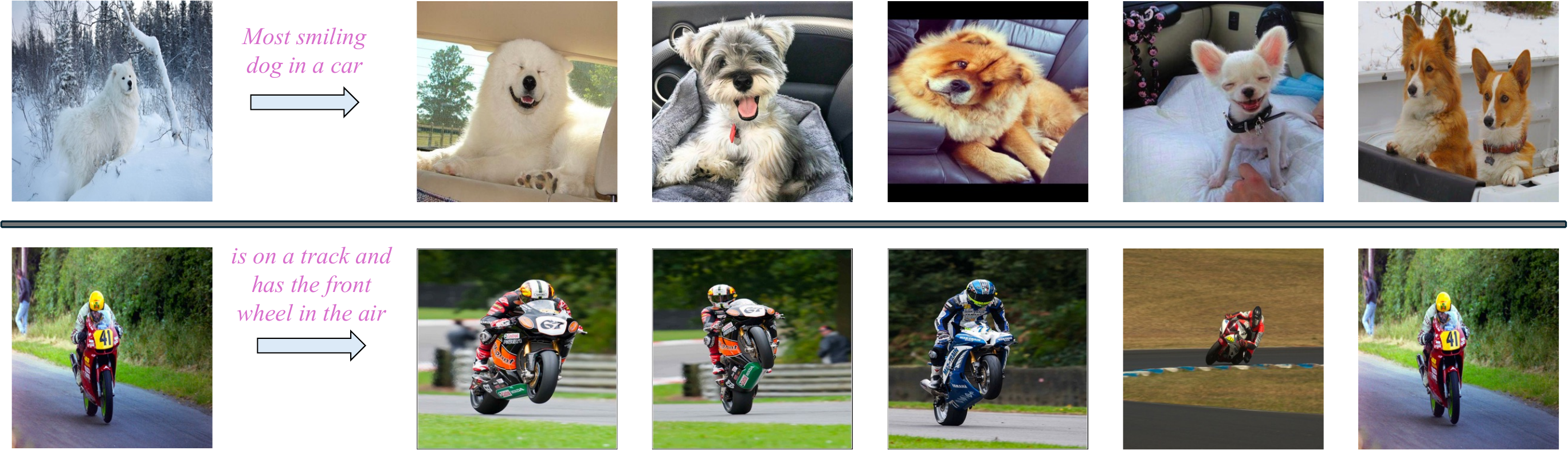}
    \caption{Examples from \textbf{CIRR} (top) and \textbf{CIRCO} (bottom) validation sets. Results are ranked from the highest (left) to lowest (right) similarity. InstructCIR effectively retrieves images across a wide variety of modifier instructions from source images.}
    \label{fig: visualized_examples}
\end{figure*}

Figure \ref{fig: visualized_examples} visualizes examples from CIRR and CIRCO with instructions impacting different semantic elements of the reference image such as viewpoint, layouts, object counts, poses, and background changes. This provides further indication about the diverse applicability of our setup.

For the FashionIQ dataset, Table \ref{tab: fiq_results} highlights the performance of our model compared to previous zero-shot methods. Our model achieves impressive improvements, with 6.04\% and 5.35\% increases in average $R@10$ over Pic2Word and SEARLE, respectively. It is important to note that our training data primarily consists of natural images, whereas FashionIQ is a domain-specific dataset focused on fashion e-commerce images. This significant performance on FashionIQ demonstrates the strong generalization capability of our model, which can effectively transfer knowledge from natural image domains to more specialized image retrieval tasks. These results illustrate the proficiency of our model in addressing the diverse challenges posed by both fashion-specific and general natural image datasets.

For the GeneCIS dataset, Table \ref{tab: genecis_results} demonstrates the superiority of our model. It surpasses Pic2Word and SEARLE by 6.85\% and 3.65\% in average $R@1$, ouerperforming all baselines in all metrics, which demonstrates its outstanding capability in processing conditional image retrieval.
\begin{table}
\centering
\scriptsize
\scalebox{1.0}{
\begin{tabular}{lccc}\toprule
\multirow{2}{*}{\multirow{2}{*}{Method}} & \multicolumn{3}{c}{\textbf{Average}} \\
\cmidrule(lr){2-4}
&$R@1$ & $R@2$ & $R@3$ \\
\midrule
Pic2Word~\citep{saito2023pic2word} &11.15 &21.47 &30.38 \\
SEARLE~\citep{baldrati2023zero} &14.35 &25.28 &34.93 \\
LinCIR~\cite{gulanguage} &12.19 &22.76 &32.38\\
CIReVL~\citep{karthik2023vision} &15.85 &27.13 & 36.33 \\
\midrule
\textbf{InstructCIR$_{lp}$} & \underline{17.37} & \underline{29.07} & \textbf{39.64} \\
\textbf{InstructCIR$_{full}$} &\textbf{18.00} &\textbf{30.07} & \textbf{39.64} \\
\bottomrule
\end{tabular}
}
\caption{\textbf{Comparison of Zero-Shot CIR Models on GeneCIS}.}
\label{tab: genecis_results}
\end{table}

\section{Ablation Study}
Our ablation studies aim to address the following key questions regarding the effectiveness and robustness of our proposed method:
\textbf{Q1:} How do different training stages contribute to model performance?
\textbf{Q2:} What is the impact of training data on model effectiveness?
\textbf{Q3:} Can our approach be easily adapted to sophisticated MLLM mechanisms?
Due to the space limitation, we defer the discussion through attention maps about how InstructCIR focuses on salient objects in Appendix \ref{sec: attn_map_analy}.

\subsection{How do different training stages contribute to model performance?}
\begin{table}[!h]
\centering
\scriptsize
\setlength{\tabcolsep}{3pt}
\begin{tabular}{cccccccc}\toprule
\multicolumn{2}{c}{Stage 1} & Stage 2 & CIRCO & CIRR & FashionIQ & \multirow{2}{*}{Avg.} \\
\cmidrule(lr){1-2} \cmidrule(lr){3-3} \cmidrule(lr){4-6}
LLaVA-Pretrain &FOIL &CC3M-Instruct &mAP@5 &R@10 &R@10 \\
\midrule
\Checkmark & \XSolidBrush & \XSolidBrush & 5.10 & 46.90 & 20.18 & 24.06 \\
\XSolidBrush & \Checkmark & \XSolidBrush & 5.22 & 47.12 & 21.35 & 24.56 \\
\Checkmark & \Checkmark & \XSolidBrush & 5.92 & 50.25 & 24.90 & 27.02 \\
\XSolidBrush & \XSolidBrush & \Checkmark & 19.43 & 73.20 & 26.66 & 39.76 \\
\Checkmark & \XSolidBrush & \Checkmark & 20.10 & 75.80 & \underline{31.73} & \underline{42.54} \\
\XSolidBrush & \Checkmark & \Checkmark & \underline{20.43} & \textbf{77.04} & 29.94 & 42.47 \\
\Checkmark & \Checkmark & \Checkmark & \textbf{21.27} & \underline{76.23} & \textbf{32.55} & \textbf{43.35} \\
\bottomrule
\end{tabular}
\caption{\textbf{Results of different stages.} LLaVA-Pretrain and/or FOIL are used in the first stage, which contain image-caption pairs. The triplet dataset CC3M-Instruct is used in the second stage. \textbf{Bold} indicates the highest scores and \underline{Underline} indicates the second highest scores.}
\label{tab: diff_stages}
\end{table}

To assess the impact of each stage in our training strategy, we conducted ablation studies, isolating the contributions of Stage 1 and Stage 2. As presented in Table \ref{tab: diff_stages}, the combination of both stages consistently yields superior performance, with the second stage contributing more significantly.
Stage 1 establishes a robust joint embedding space for images and text through contrastive learning on image-caption pairs. Though not directly related to CIR, it reduces the modality gap, which is crucial for handling complex compositional queries in Stage 2. In contrast, Stage 2 directly aligns the model's training objective with the CIR task by using triplet-based contrastive learning. Here, the model is explicitly trained to match the image-modification pair to the modified caption, which mirrors the actual CIR task during inference. This stage fine-tunes the model to follow modification instructions and adapt its embeddings accordingly. By directly optimizing for the target task, Stage 2 has a more substantial influence on final performance. We observe that Stage 2 alone, without the pre-alignment from Stage 1, performs suboptimally, indicating that the initial feature alignment plays a critical supporting role. This interplay between stages highlights the importance of a progressive learning strategy that first handles modality discrepancies before transitioning to task-specific fine-tuning. Additionally, the combination of LLaVA-Pretrain and FOIL in the first stage performs better than using either dataset alone, emphasizing the importance of exposing the model to diverse data during feature alignment and the effectiveness of our training strategy. 

\subsection{What is the impact of training data on model effectiveness?}
\begin{figure}
    \centering
    \includegraphics[width=0.8\linewidth]{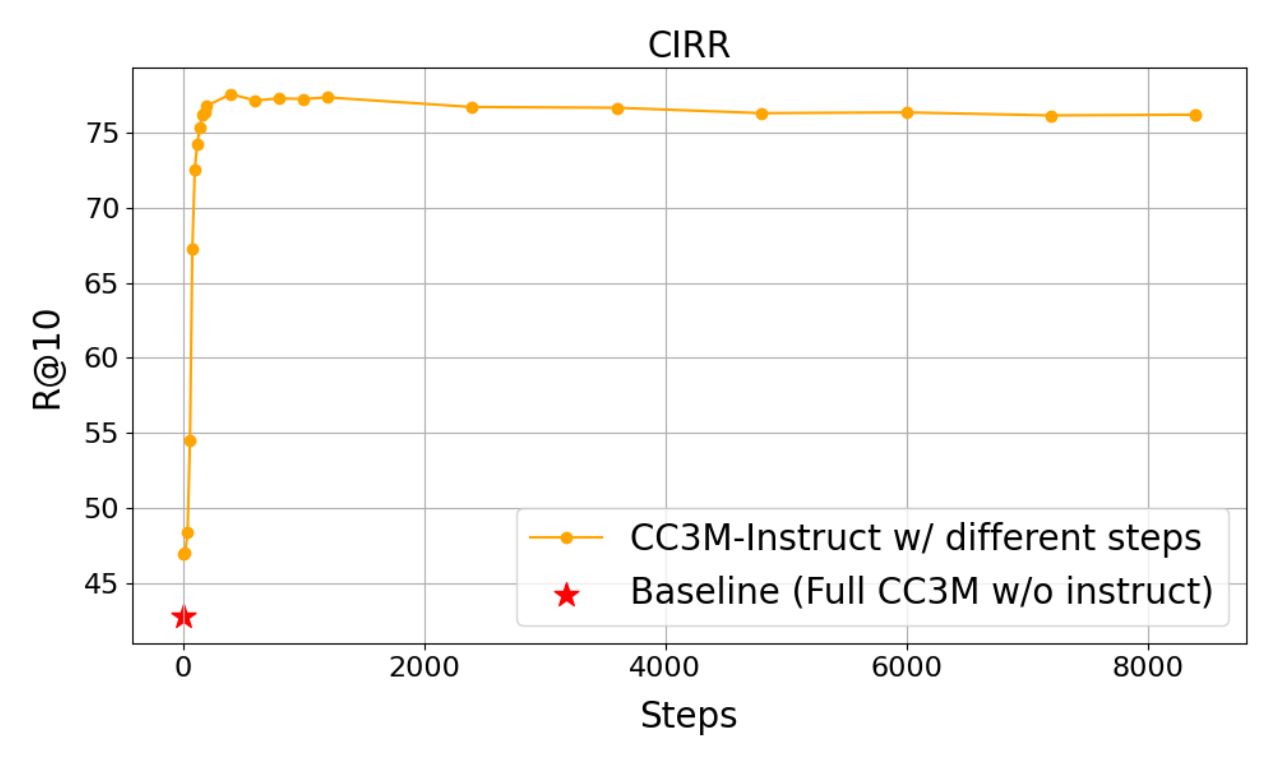}
    \caption{\textbf{Effectiveness of the triplet data by scale.} The baseline is our model trained with the whole original CC3M pair data. The plot demonstrates the performance curve on validation sets by steps. The performance improves rapidly at beginning steps.}
    \label{fig: data_effect}
\end{figure}

To evaluate the effectiveness of the triplet dataset and the training scale, we conducted experiments using different dataset sizes of the CC3M-Instruct and the original pair-wise CC3M datasets. Figure \ref{fig: data_effect} shows the performance across varying training steps. We observe that using the entire original pair data yields results similar to those obtained in the first-stage datasets, whereas the use of triplet data significantly improves performance.
The recall grows rapidly up to around 1200 steps (approximately 300K triplets), after which it stabilizes. Continued training introduces fluctuations and potential overfitting. This pattern suggests that MLLMs quickly adapt to the training data, emphasizing the importance of carefully managing training data scale. We find that a 300K subset balances efficiency and performance, and recommend using diverse, regular-sized datasets for future training.

\begin{table}[!htp]\centering
\scriptsize
\begin{tabular}{lcccccc}\toprule
\multirow{2}{*}{\multirow{2}{*}{\shortstack{Hard Neg. \& \\ Template Strategy}}} &\multicolumn{3}{c}{CIRR} &\multicolumn{2}{c}{FashionIQ} \\
\cmidrule(lr){2-4} \cmidrule(lr){5-6}
&$R@1$ &$R@5$ &$R@10$ &$R@10$ &$R@50$ \\
\midrule
$(i, c)$, $c_r$ as hard neg. & 10.79 & 33.10 & 46.04 & 9.24 & 21.28 \\
$(i, t, c_r)$ w/o hard neg. & 33.08 & 63.98 & 75.58 & 30.96 & 51.05 \\
Fixed templates & 33.60 & 63.81 & 74.85 & 31.32 & 52.06 \\
\midrule
Ours & \textbf{34.63} & \textbf{64.90} & \textbf{76.23} & \textbf{32.55} & \textbf{52.54} \\
\bottomrule
\end{tabular}
\caption{\textbf{Results of Different Hard Negative and Template Strategies.} “Ours” denotes the use of $(i, t, c_r)$ triplets with $c$ as hard negatives and randomly selected templates during training, as opposed to fixed templates. 
}
\label{tab: diff_hard_negs}
\end{table}
In the second stage of training, we use the original caption $c$ as the hard negative of a triplet $(i, t, c_r)$. In Table \ref{tab: diff_hard_negs}, we show the effectiveness of incorporating hard negatives. It can be observed that the incorporation of hard negatives improve the performance because the modified caption and original caption may look similar and contrasting them in training can enhance the model ability to understand the difference. In addition, the first row in the table shows the opposite strategy that uses the original image-caption pair with the modified caption as the hard negative. Results again signify the effectiveness of training with the modification query and modified caption against the original pair. Furthermore, in the second training stage, we utilize randomly selected prompt templates, whereas row 3 demonstrates the opposite approach by using fixed prompts for training. The results reveal the necessity of employing diverse templates.

\subsection{Can our approach be easily adapted to sophisticated MLLM mechanisms?}
\label{sec: advanced_mllm}
In this section, we analyze sophisticated MLLM mechanisms with a latest MLLM microsoft/Phi-3.5-vision-instruct on our training strategies. The difference between microsoft/Phi-3.5-vision-instruct and LLaVA-Phi are two folds: (1) The former is trained with three stages including the feature alignment, instruction tuning, and preference optimization~\citep{rafailov2024direct} while the latter is only trained with the first two stages; (2) Phi-3.5-vision-instruct leverages the dynamic high resolution~\citep{liu2024llavanext, dong2024internlm}. An input image that is oversize will not only be resized but also chunked into several parts. The resized image and image parts will be encoded by the visual encoder and fed to the LLM together. While such an operation is powerful, it also suffers from higher computational cost in both training and inference as more patches are fed to the LLM.
\begin{table}
\scriptsize
\centering
\begin{tabular}{lccccccc}
\toprule
\multirow{3}{*}{Method} & \multicolumn{6}{c}{\textbf{CIRR}} \\ 
\cmidrule(lr){2-7}
& $R@1$   & $R@5$   & $R@10$ & $R_s @1$ & $R_s @2$ & $R_s @3$ \\
\midrule
E5-V & 34.17 & 64.39 & 75.98 & 64.02 & 83.35 & 92.70 \\
\textbf{InstructCIR} & 35.18 & 65.12 & 77.61 & 67.54 & 84.77 & 93.61 \\
\textbf{InstructCIR+} & \textbf{37.93} & \textbf{69.40} & \textbf{80.36} & \textbf{70.72} & \textbf{87.18} & \textbf{94.34}\\
\bottomrule
\end{tabular}
\caption{\textbf{Results of sophisticated MLLMs on CIRR Test Set}. InstructCIR uses LLaVA-Phi as the base model, consistent with the main experiments, while InstructCIR+ uses microsoft/Phi-3.5-vision-instruct as the base model.}
\label{tab: phi35_vision_cirr}
\end{table}
\begin{table}[!ht]
\scriptsize
\centering
\begin{tabular}{lcccccccccc}
\toprule
\multirow{3}{*}{Method} & \multicolumn{4}{c}{\textbf{CIRCO}} \\ 
\cmidrule(lr){2-5}
& $mAP@5$ & $mAP@10$ & $mAP@25$ & $mAP@50$ \\
\midrule
E5-V & 20.44 & 22.06 & 24.25 & 25.26 \\
\textbf{InstructCIR} & 22.32 & 23.80   & 26.25  & 27.32 \\
\textbf{InstructCIR+} & \textbf{26.12} & \textbf{27.18} & \textbf{29.50} & \textbf{30.53} \\
\bottomrule
\end{tabular}
\caption{\textbf{Results of sophisticated MLLMs on CIRCO Test Set}.}
\label{tab: phi35_vision_circo}
\end{table}



We use the microsoft/Phi-3.5-vision-instruct model as the base to conduct ablations on the CIRR and CIRCO \textit{test sets}, referring to this variant as \textbf{InstructCIR+}. We compare it with a concurrent work, E5-V~\citep{jiang2024e5}, which utilizes LLaVA-NeXT~\citep{liu2024llavanext, liu2023improvedllava} as the backbone—a twice larger MLLM equipped with dynamic high resolution. Our method differs from E5-V in that our training strategy is multimodal and instruction-aware, whereas E5-V trains the MLLM only on pure text pair data. Results are shown in Table \ref{tab: phi35_vision_cirr} and \ref{tab: phi35_vision_circo}. As observed, Phi-3.5-Vision improves upon LLaVA-Phi despite both using Phi-3.5-mini as LLMs. These findings indicate that these techniques can benefit CIR and that our training strategy can be directly applied to existing MLLMs. Notably, both InstructCIR and InstructCIR+ outperform E5-V, even without LLaVA-Phi using dynamic high resolution, highlighting the effectiveness of our instruction-aware training strategy.
\section{Conclusion}
In this paper, we present InstructCIR, a ZS-CIR model built on instruction-tuned MLLMs. Our approach highlights the potential of MLLMs in CIR systems, leveraging their robust instruction-following abilities and strong vision-language alignment to address the lack of instruction-awareness in previous methods. The proposed two-stage training strategy effectively refines the MLLM’s text generation capabilities for embedding extraction while enhancing its instruction-following within the CIR context. We believe this work provides valuable insights into model selection and training strategies, paving the way for future advancements in Composed Image Retrieval.
{
    \small
    \bibliographystyle{ieeenat_fullname}
    \bibliography{main}
}
\clearpage
\setcounter{page}{1}

\maketitlesupplementary
\renewcommand{\thesection}{\Alph{section}}
\setcounter{section}{0}

\section{Related Works}
\subsection{Instruction Tuning}
Instruction tuning~\citep{zhang2023instruction, ouyang2022training, chung2024scaling, zheng2023judging} is a strategy commonly adopted in modern LLM training to enhance model generalization by exposing models to various prompts. In the realm of multimodal large language models (MLLMs), visual instruction tuning~\citep{liu2024visual} has significantly improved their instruction-following capabilities when processing multimodal data. This process typically involves two stages: the first stage trains an adapter between the visual encoder and the LLM using image captioning data; in the second stage, the LLM and the adapter are jointly trained with instruction-following data that encompasses multiple tasks in a question-answer format. While previous MLLMs have primarily focused on text generation, recent research is exploring the use of LLMs for representation learning. Specifically, E5-Mistral~\citep{wang2023improving} leverages LLMs as embedding models by training them on various retrieval tasks specified by instructions. E5-V~\citep{jiang2024e5} extends this approach to multimodal domains; however, its training remains based on pure text pairs, and the full potential of MLLMs for multimodal embeddings is not fully realized. In this paper, we propose a novel approach to train an instruction-aware model that generates multimodal embeddings through two stages: embedding alignment and instruction contrastive learning.

\subsection{Composed Image Retrieval}
Composed Image Retrieval (CIR) involves finding images related to a source image under a specified condition, typically provided as a modifier text. This task has practical applications in e-commerce, recommendation systems, and more. Due to the difficulty of acquiring specific datasets for various CIR tasks, recent research has focused on Zero-Shot CIR (ZS-CIR). Previous methods primarily represent the reference image as specific tokens and concatenate them with text tokens for retrieval~\citep{saito2023pic2word, karthik2023vision, tang2024context, suo2024knowledge, agnolucci2024isearle,gulanguage}. With the advent of Multimodal Large Language Models (MLLMs), researchers have begun incorporating LLMs into this domain. For instance, CIReVL~\citep{karthik2023vision} leverages two MLLMs: one for generating image captions and another for combining captions with modifier texts for retrieval. FROMAGe~\citep{koh2023grounding} and MCL~\citep{liimproving} explore using LLMs for embeddings, but the LLMs are mainly used as text encoders. Despite the rapid development of MLLMs exhibiting strong generalization, instruction-following, and zero-shot capabilities in multimodal data, their applications to CIR tasks are rarely explored. In this paper, we leverage MLLMs as embedding models for CIR tasks, enabling direct encoding of images and modifier texts within a single model.
\begin{figure*}[h]
    \centering
    \includegraphics[width=0.7\linewidth]{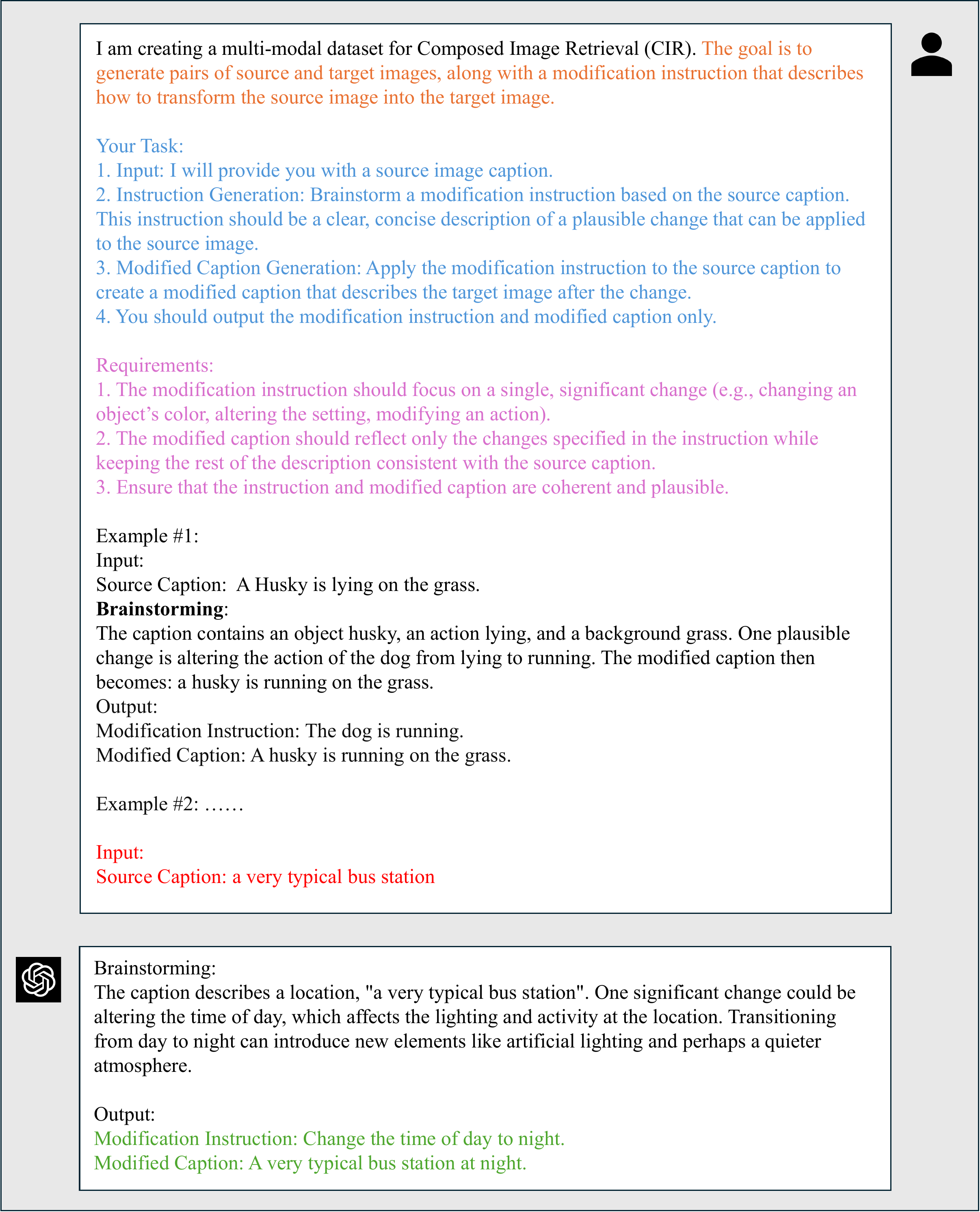}
    \caption{We prompt GPT-4o to generate triplet data from CC3M. Our prompt consists of three parts: the first part (\textcolor{orange}{orange}) defines the task we aim to complete; the second part (\textcolor{myblue}{blue} and \textcolor{mypurple}{purple}) specifies the details and requirements of the task; and the third part (\textcolor{black}{black}) provides examples for triplet generation, where the modifier text is brainstormed step by step. The key concepts in the captioned are identified and subsequently selected concepts are altered. The modified caption is derived accordingly. Finally, we provide the input (\textcolor{red}{red}). GPT then outputs the modifier text and the corresponding caption based on the query caption (\textcolor{mygreen}{green}).}
    \label{fig: gpt_prompt}
\end{figure*}

\section{Dataset Details}
\label{sec: data_detail}
We evaluate our model using four well-established zero-shot CIR benchmarks: FashionIQ~\citep{guo2019fashion}, CIRR~\citep{Liu_2021_ICCV}, CIRCO~\citep{baldrati2023zeroshot}, and GeneCIS~\citep{vaze2023genecis}. While FashionIQ is an early benchmark for CIR, its domain is restricted to fashion e-commerce images. In contrast, CIRR and CIRCO focus on more general natural images. CIRR is the first CIR dataset centered on natural images, but it suffers from the limitation of having only one target image per query, leading to potential false negatives. On the other hand, CIRCO improves upon this by providing multiple target images per query, which reduces the likelihood of false negatives and offers a more comprehensive evaluation of retrieval accuracy. 
GeneCIS is a dataset for conditional image retrieval. It defines four types of conditions as focusing or changing attributes or objects in images. In line with common practice, we report Recall@k ($R@k$) for FashionIQ, CIRR, and GeneCIS, with an additional subset metric for CIRR denoted as $R_s@k$. For CIRCO, where multiple correct images can correspond to a single query, we use mean Average Precision ($mAP@k$) to capture both precision and recall across different retrieval positions. 

\section{Triplet Data Generation}
\label{sec: triplet_data}

\subsection{Data Processing}
We utilize GPT-4o~\citep{achiam2023gpt} to process and generate triplet data. Given an image and its caption, we use the caption as a prompt to GPT, which then derives the modifier text and the modified caption. The detailed prompt structure is shown in Figure \ref{fig: gpt_prompt}. Specifically, the prompt is divided into three parts: task definition, requirements, and few-shot examples.

Our data generation process differs from MCL~\citep{liimproving} in several aspects. First, we leverage GPT-4o~\citep{achiam2023gpt} instead of LLAMA2~\citep{touvron2023llama}, allowing for more generalizable and creative content generation. Second, GPT-4o has a larger context window, enabling us to incorporate more complex techniques within the prompt. Unlike MCL, which directly presents the output modifier text and corresponding caption in few-shot examples, we divide the generation process into several steps using the Chain of Thought method~\citep{wei2022chain}. We instruct GPT to first identify key points in the example caption, then selectively alter some of them as modifications, and finally derive the modified caption. This step-by-step generation ensures that the generated modifier text and corresponding caption are reasonable and closely related to the original caption. \textit{At the time the major work of this paper is finished, the MCL dataset has not been released. We will defer the comparison between two datasets in the future work.}

Our pipeline differs from the training set derivation in \citep{vaze2023genecis}. While they use text scene graphs to identify subjects, predicates, and objects, their modifier instruction is generated by simply replacing one element with another concept from the dataset, leading to limited creativity and diversity.

\begin{table}
\centering
\begin{tabular}{c|c}
\hline
\textbf{Dataset} & \textbf{Approx. Size} \\ \hline
FOIL~\cite{shekhar2017foil_acl} & 60K  \\
LLaVA-Pretrain~\cite{liu2024visual} & 558K \\
CC3M-Instruct & 300K \\
\hline
\end{tabular}
\caption{Approximate Sizes of Different Datasets}
\label{tab: dataset_size}
\end{table}

\begin{table*}[!h]
\centering
\begin{tabular}{l|p{11cm}}
\hline
Task & Instruction Template \\
\hline
\multirow{13}{*}{\shortstack{Image \\ Modification}} & \texttt{<Image>} The image is conditioned on the following prompt: \{modifier text\}, summarize the image and the prompt to retrieve a description of the image changed by the condition: \\
                                   & \texttt{<Image>} Given the image conditioned by the prompt: \{modifier text\}, condense the essence of the image and the prompt into a single word to fetch a description of the altered image: \\
                                   & \texttt{<Image>} Using the prompt to condition the image: \{modifier text\}, provide one word that encapsulates the overall concept of the conditioned image to retrieve its description: \\
                                   & \texttt{<Image>} Based on the image influenced by this prompt: \{modifier text\}, distill the description of the conditioned image and the prompt into one word to access the altered description: \\
                                   & \texttt{<Image>} With the image modified according to the prompt: \{modifier text\}, summarize both the image and the prompt to obtain a description of the conditioned image: \\
                                   & \texttt{<Image>} Condition the image with this condition: \{modifier text\}. Summarize the result: \\
                                   & \texttt{<Image>} Using this prompt: \{modifier text\}, describe the conditioned image: \\
                                   & \texttt{<Image>} Apply the prompt: \{modifier text\} to the image. Provide one word for the conditioned image: \\
                                   & \texttt{<Image>} Given this prompt: \{modifier text\}, condense the conditioned image into one word: \\
                                   & \texttt{<Image>} \{modifier text\}: \\ \hline
\multirow{6}{*}{\shortstack{Caption \\ Summary}} & \texttt{<Caption>} Summary: \\
                     & \texttt{<Caption>} Summarize the caption for retrieval: \\
                     & \texttt{<Caption>} A shorter description is: \\
                     & \texttt{<Caption>} Shorter caption: \\
                     & \texttt{<Caption>} “”\\ \hline
\end{tabular}
\caption{Instruction templates for different tasks. In \textbf{Image Modification}, the modifier text combined with the selected template serves as the formatted prompt. \textbf{Caption} \textbf{Summary} instruct the model to generate a global representation for captions.}
\label{tab: prompt_template}
\end{table*}

\subsection{Data Details}
\label{sec: data_stats}
\begin{figure*}
    \centering
    \includegraphics[width=0.7\linewidth]{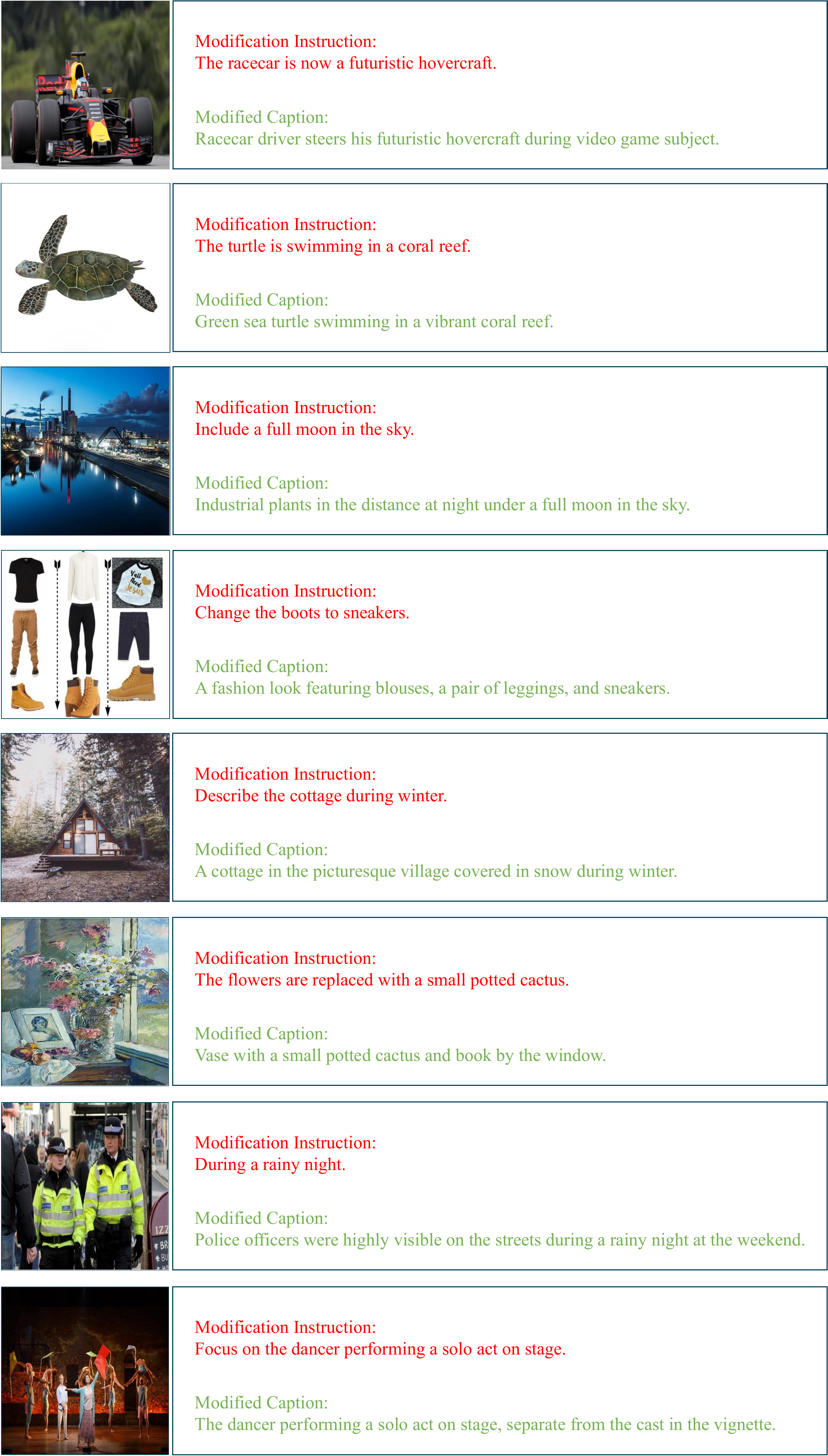}
    \caption{Triplet Examples from CC3M-Instruct}
    \label{fig: vis_example}
\end{figure*}

After filtering invalid images and failed prompts, we acquire the CC3M-Instruct dataset with 2M triplets. We randomly sample 300K triplets to maintain the training efficiency as well as performance. Triplet examples are shown in Figure \ref{fig: vis_example}. In our experiments, the first stage of training takes about 1.5 hours while the second stage of training takes about 2.5 hours. Table \ref{tab: dataset_size} shows the sizes of training datasets. 

\section{Prompt Templates}
\label{sec: prompt_tempalate}
Templates for training are shown in Table \ref{tab: prompt_template}.

\subsection{Templates for Training}
In our second training stage, we use two prompt templates for the source image and modification instruction, and the target caption. Both templates are sampled from predefined template sets, respectively. Table \ref{tab: prompt_template} shows both template sets we used.

\subsection{Templates for Zero-shot Inference}
We use different prompt template for inference. In inference, the source image and modification instruction are formatted in a prompt. The target image are composed with a summary prompt. Both are encoded by InstructCIR and a consine similarity is computed between their embeddings.

\textbf{CIRR \& CIRCO}

\textit{Image Captioning}

\texttt{<Image>} Describe this image in one word: 

\textit{Image Modification}

\texttt{<Image>} Modify this image with \{modifier text\}, describe the modified image in one word: 

\textbf{FashionIQ}

\textit{Image Captioning}

\texttt{<Image>} Describe this \{data type in fashioniq\} in one word based on its style: 

\textit{Image Modification}

\texttt{<Image>} Modify the style of this \{data type in fashioniq\} based on \{modifier text\}. describe this modified \{data type in fashioniq\} in one word based on its style: 

\textbf{GeneCIS}

\textit{Image Captioning}

\texttt{<Image>} Summarize the image for retrieval: 

\textit{Image Modification}

\texttt{<Image>} Describe the image in one word with a specific focus on the attribute \{specific attribute\}:

\texttt{<Image>} Describe the image in one word with a specific change of the attribute \{specific attribute\}:

\texttt{<Image>} Describe the image in one word with a specific focus on the object \{specific object\}:

\texttt{<Image>} Describe the image in one word with a specific change of the object \{specific object\}:

\section{Attention Map Analysis}
\label{sec: attn_map_analy}
In this section, we analyze what InstructCIR learns for composed image retrieval. Specifically, we aim to investigate which parts of the original image contribute the most to the composed embedding. Note that the composed embedding contains both the image and instruction. Therefore, the most significant parts are supposed to be indicated by the instruction instead of just the major ones. Inspired by \cite{yu2024attention} that creates attention maps highlighting instruction-aware image patches, we conduct qualitative analysis through attention maps. Specifically, InstructCIR leverages the [EOS] token embedding from the output sequence as the composed embedding $h_{it}$. Similar to \cite{yu2024attention}, we compute the similarity between the composed embedding $h_{it}$ and patch embeddings $H=\{h_1, h_2,\cdots,h_{P^2}\}$ in the output sequence, where $P$ is the number of image patches. The patch similarity is resized to the grid shape $S \in \mathbb{R}^{P\times P}$. The similarity grid is finally interpolated to the attention map $A \in \mathbb{R}^{H\times W}$, where $H$ and $W$ are the height and width of the image. 

\begin{figure}
    \centering
    \includegraphics[width=1.0\linewidth]{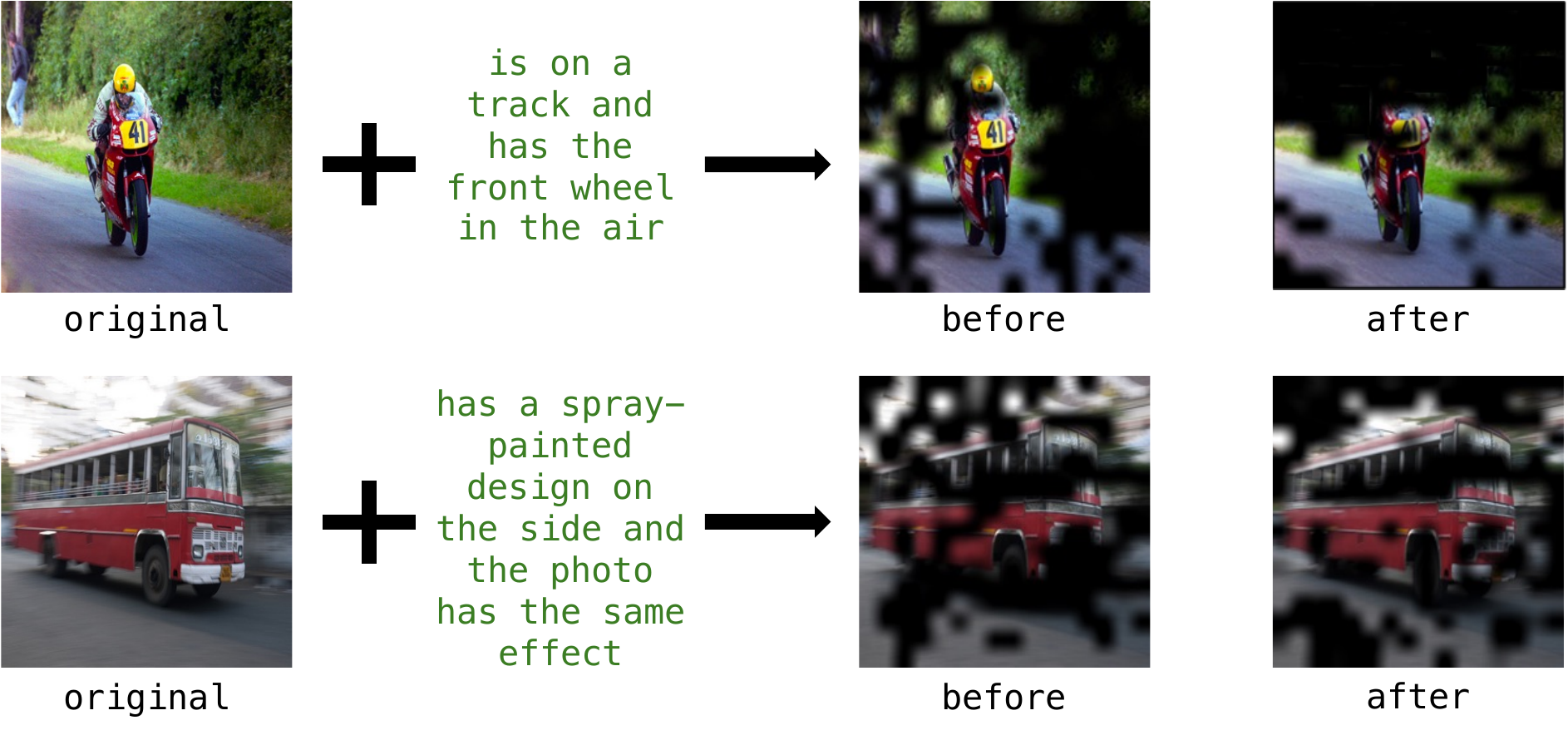}
    \caption{\textbf{Qualitative Examples of Attention Maps.} Before training, the model can only or cannot focus on specific parts in the image according to the instruction. After the training, the model is able to capture these parts. In the example above, the model highlights the front wheel and floor. In the example below, the model high lights the main side of the bus.}
    \label{fig: attn_exam}
\end{figure}

\begin{equation}
\begin{aligned}
    S &= H \cdot (h_{it})^T, S\in \mathbb{R}^{P^2} \\
    S &= \text{Resize}(S), S\in \mathbb{R}^{P\times P} \\
    A &= \text{Interpolate}(S), A \in \mathbb{R}^{H\times W}
\end{aligned}
\label{equ: attn_cal}
\end{equation}

The attention map is mixed with the image in the alpha channel, highlighting important objects in the image. Examples are shown in the Figure \ref{fig: attn_exam}, demonstrating the original image, attention maps before and after training the model. 
Before training, the original model (LLaVA-Phi) focuses on the main object in the image. In the first example, it mainly focuses on the motorman. After training, the model is able to pay higher attention on specific parts mentioned in the image, i.e., the front wheel and the floor, than other parts. The qualitative analysis highlights the instruction-awareness of the model trained with the two-stage strategy. 

\begin{table*}[ht]
\centering
\scriptsize
\begin{tabular}{cccccccccccccc}\toprule
\multirow{2}{*}{\multirow{2}{*}{Method}} &\multicolumn{3}{c}{\textbf{Focus Attribute}} &\multicolumn{3}{c}{\textbf{Change Attribute}} &\multicolumn{3}{c}{\textbf{Focus Object}} &\multicolumn{3}{c}{\textbf{Change Object}} & \textbf{Average} \\
\cmidrule(lr){2-4} \cmidrule(lr){5-7} \cmidrule(lr){8-10} \cmidrule(lr){11-13} \cmidrule(lr){14-14}
&$R@1$ &$R@2$ &$R@3$ &$R@1$ &$R@2$ &$R@3$ &$R@1$ &$R@2$ &$R@3$ &$R@1$ &$R@2$ &$R@3$ & $R@1$ \\
\midrule
Pic2Word &15.65 &28.16 &38.65 &13.87 &24.67 &33.05 &8.42 &18.01 &25.77 &6.68 &15.05 &24.03 &11.15 \\
SEARLE &17.10 &29.60 &40.70 &\textbf{16.30} &25.20 &34.20 &12.00 &22.20 &30.90 &12.00 &24.10 &33.90 &14.35 \\
LinCIR &16.90 &29.95 &41.45 &16.19 &27.98 &36.84 &8.27 &17.40 &26.22 &7.40 &15.71 &25.00 &12.19\\
CIReVL &19.50 &31.80 &42.00 &14.40 &26.00 &35.20 &12.30 &21.80 &30.50 &\textbf{17.20} &28.90 &37.60 &15.85 \\
\midrule
\textbf{InstructCIR} &\textbf{21.25} &\textbf{34.55} &\textbf{46.85} &16.15 &\textbf{28.74} &\textbf{39.73} &\textbf{17.55} &\textbf{28.01} &\textbf{36.94} &17.04 &\textbf{28.98} &\textbf{37.70} &\textbf{18.00} \\
\bottomrule
\end{tabular}
\caption{\textbf{Comparison of Zero-Shot CIR Models on GeneCIS}.}
\label{tab: complete_genecis_results}
\end{table*}

\begin{table}[ht]
\centering
\begin{tabular}{c|c}
\hline
\textbf{Model} & \textbf{Size} \\ \hline
xtuner/llava-phi-3-mini-hf & 4.14B  \\
microsoft/Phi-3.5-vision-instruct & 4.15B \\
E5-V & 8.35B \\
\hline
\end{tabular}
\caption{Number of parameters of different models}
\label{tab: model_size}
\end{table}

\subsection{InstructCIR Training}
Detailed training configs are shown in Table \ref{tab: icir_config}.

\begin{table}[ht]
\centering
\begin{tabular}{c|c}
\hline
\textbf{Training Config} & \textbf{Value} \\ \hline
DeepSpeed & ZeRO-2 \\ 
LoRA R & 64 \\
LoRA Alpha & 16 \\
Model Max Length & 512 \\
Precision & FP16 \\
Epochs for both stages & 1 \\
Batch Size Per GPU in Stage 1 & 48 \\
Batch Size Per GPU in Stage 2 & 64 \\
Gradient Accumulation Steps & 1 \\
Learning Rate & 2E-05 \\
Weight Decay & 0 \\
Warm Up Ratio & 0.03 \\
LR Scheduler Type & Cosine \\ \hline
\end{tabular}
\caption{Configurations of Training InstructCIR.}
\label{tab: icir_config}
\end{table}

\begin{table}[ht]
\centering
\begin{tabular}{c|c}
\hline
\textbf{Config} & \textbf{Value} \\ \hline
Visual Encoder & openai/clip-vit-large-patch14 \\
Image Resolution & 224x224 \\
Language Model & microsoft/Phi-3.5-mini-instruct \\
Adapter & MLP \\
Pretraining Strategy & Frozen LLM, Frozen ViT \\
Fine-tuning Strategy & Full LLM, Full ViT \\
Pretrain Dataset & ShareGPT4V-PT (1246K)~\citep{chen2023sharegpt4v} \\
Fine-tune Dataset & InternVL-SFT (1268K)~\citep{chen2024internvl} \\
Pretrain Epoch & 1 \\
Fine-tune Epoch & 2 \\
\hline
\end{tabular}
\caption{Configurations of Training LLaVA-Phi}
\label{tab: mllm_config}
\end{table}

\section{Training Details}
\label{sec: train_details}
\subsection{MLLM Training}
We use the code and data from xtuner/llava-phi-3-mini-hf~\citep{2023xtuner} to train a variant of LLaVA-Phi. \textit{Note that the goal of this step is solely to make our experiments consistent with the baselines.} Section 4.3 has demonstrated that our training strategy can be directly applied to existing MLLMs. The checkpoint of the variant LLaVA-Phi will also be released for reproducibility. MLLM training and model details are provided in Table \ref{tab: mllm_config}. In addition, InstructCIR and InstructCIR+ leverage on-device MLLMs which are much smaller than the current work E5-V~\cite{jiang2024e5}. Their numbers of parameters are shown in Table \ref{tab: model_size}.

\section{More Experiment Results}
\label{sec: more_results}
Table \ref{tab: complete_genecis_results} shows the complete results on GeneCIS.

\end{document}